\documentclass[letterpaper, 10 pt, conference]{ieeeconf}  

\IEEEoverridecommandlockouts                              

\usepackage{amsmath} 
\usepackage{amssymb}  
\usepackage{graphicx}
\usepackage{xcolor}
\usepackage{color}
\usepackage{hyperref}

\title{\LARGE \bf
Koopman Operator Based Linear Model Predictive Control for 2D Quadruped Trotting, Bounding, and Gait Transition
}

\author{Chun-Ming Yang and Pranav A. Bhounsule
\thanks{Dept. of Mechanical and Industrial Engineering, University of Illinois at Chicago, 
       842 W Taylor St, Chicago, IL 60607, USA. Email:
       {\tt\small jyang241@uic.edu}, {\tt\small pranav@uic.edu}
       The work was supported by NSF grant 2128568.}%
}

\begin{document}

\maketitle
\thispagestyle{empty}
\pagestyle{empty}

\begin{abstract}
Online optimal control of quadrupedal robots would enable them to plan their movement in novel scenarios. Linear Model Predictive Control (LMPC) has emerged as a practical approach for real-time control. In LMPC, an optimization problem with a quadratic cost and linear constraints is formulated over a finite horizon and solved on the fly. However, LMPC relies on linearizing the equations of motion (EOM), which may lead to poor solution quality. In this paper, we use Koopman operator theory and the Extended Dynamic Mode Decomposition (EDMD) to create a linear model of the system in high dimensional space, thus retaining the nonlinearity of the EOM. We model the aerial phase and ground contact phases using different linear models. Then, using LMPC, we demonstrate bounding, trotting, and bound-to-trot and trot-to-bound gait transitions in level and rough terrains. The main novelty is the use of Koopman operator theory to create hybrid models of a quadrupedal system and demonstrate the online generation of multiple gaits and gaits transitions.
\end{abstract}

\section{Introduction}

Among legged systems, quadrupedal robots have emerged as a ubiquitous platform for designing and testing radically novel control ideas. A dominant approach is to perform online optimal control using a linearized model (e.g. Model Predictive Control) to enable real-time adaptation. However, the linearization may introduce model artifacts that could lead to poor performance. This paper addresses the limitation by using Koopman operator theory to build linear models in high dimensional space that preserves the system's nonlinearity. Furthermore, using Linear Model Predictive Control, we demonstrate multiple gaits (bounding and trotting) and gaits transitions. 

%
%

The earliest work investigated simple heuristics, such as regulating foot placement for speed control and vertical force for height control, and applied it to bounding, trotting, and pronking for a quadruped \cite{raibert1990trotting}. However, such controllers need to be manually tuned to be robust, which is often time-consuming and not generalizable to other systems.  
%
A formal method is to compute the control necessary to generate periodic gait, also known as the Poincar\'e limit cycle \cite{strogatz1994nonlinear}, and use eigenvalues to determine stability \cite{park2014quadruped}. Such stability analyses have limited utility because they only consider small perturbations along the periodic gaits.
%

The development of effective optimization tools has led to optimal control using Model Predictive Control (MPC). 
For instance, a quadruped bounding gait was achieved by deriving control inputs from a hierarchical nonlinear MPC  \cite{li2021model}. However, this was limited to offline computation due to the high computational time needed for the optimization.
%
Online optimal control can be achieved by solving a short-horizon optimal control problem with a quadratic cost and linear constraints, also known as Linear Model Predictive Control (LMPC) \cite{di2018dynamic}. Here, the speed-up is achieved because the optimization is a convex quadratic program, which leads to an analytical solution. However, simplicity comes at a cost: the linear constraints are obtained by linearizing the equation of motion, which may lead to poor convergence when the system deviates far from the linearization. Another issue is that such optimization assumes a pre-determined timing, which may not hold for certain gaits (e.g. bounding gait) or  rough terrain \cite{ding2019real}.
%
%
A solution to this issue is to add gaits and foothold timing as optimization variables to achieve versatile quadruped gaits, where a single optimization process plans the gait sequence, step timings, and footholds \cite{winkler2018gait}. But such optimizations are challenging to solve in real-time.

%
Deep Reinforcement Learning (DRL) is a model-free method that solves an optimization problem over the entire state space. Here, using an offline optimization, a neural network learns the mapping between sensors and actuator commands. Then during online deployment, the neural network outputs an actuator command based on the sensor value.
DRL has been demonstrated on trotting \cite{chen2023learning}, bounding \cite{bellegarda2021robust}, pacing \cite{seyfarth2003swing}, pronking \cite{gangapurwala2020guided}, and the transitions between each gait \cite{zhang2024learning}. 
However, DRL is not sample efficient as it requires an extensive dataset for training. To achieve sample efficiency, one could use model-based optimization to plan a reference motion and model-free RL to track the reference motion \cite{jenelten2024dtc}.

\begin{figure} [tbp]
\centering
\includegraphics[scale=0.9]{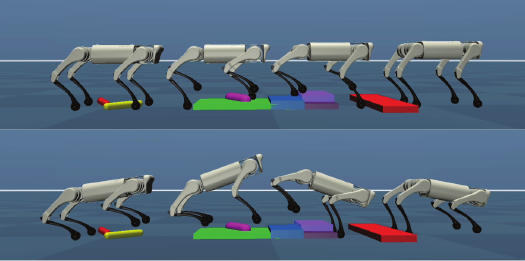}
\caption{Trotting (top) and bounding (bottom) on rough terrain. Video of all experimental results: \url{https://youtu.be/GSJi2JVUUNU}}
\label{fig:cover_pic}
\vspace{-0.5cm}
\end{figure}


The Koopman operator takes a nonlinear model $\mathbf{x}_{t+1}= \mathbf{f}(\mathbf{x}_{t}, \mathbf{u}_{t} )$ and converts it into a linear model in high dimensional space: $\mathbf{\Pi}(\mathbf{x}_{t+1})=\mathbf{A}\mathbf{\Pi}(\mathbf{x}_{t})+\mathbf{B}\mathbf{u}_{t}$, where ${\bf A}, {\bf B}$ are constant matrices and ${\bf \Pi({\bf x})}$ is a non-linear function of ${\bf x}$. The original method was conceptualized for an uncontrolled system in 1931 by Koopman \cite{koopman1931hamiltonian}, but only recently tools have been devised to model controlled systems \cite{williams2015data}. The applications of the Koopman operator are currently limited to a few simple smooth systems, such as quadcopters \cite{narayanan2023se}, underwater vehicles \cite{rahmani2024enhanced}, autonomous cars \cite{kim2001model}, two-link planar manipulators \cite{shi2021acd}, and soft robot manipulators \cite{bruder2020data}. 

In this paper, we use the Koopman operator to create multiple linear models that capture the aerial phase and ground contact phases. Although the models are linear, they capture the non-linearity by projecting in a high-dimensional space. Next, we use Linear Model Predictive Control (LMPC) to demonstrate bounding and trotting gaits and bound-to-trot and trot-to-bound gait transitions. Furthermore, we demonstrate gait robustness by introducing rough terrain. The main novelty is the use of Koopman operator theory to model and control hybrid systems such as a quadrupedal robot.

\begin{figure} [tbp]
\hspace*{0.7cm}
\includegraphics[scale=1.8]{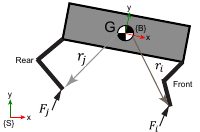}
\caption{2D Single Rigid Body model}
\label{fig:SRB}
\vspace{-0.5cm}
\end{figure}
\section{Methods}

\subsection{Quadruped Gaits Pattern}
Trotting and bounding gaits are simple in that the  Front Right (FR), Front Left (FL), Rear Right (RR), and Rear Left (RL) legs are used in pairs. In trotting, the legs move in diagonal pairs FR-RL and FL-RR, alternating between swing and stance phases. In bounding, the front legs FR-FL and rear legs RL-RR move as pairs, transitioning through a periodic cycle that includes front stance, flight phase, rear stance, and flight phase.

\subsection{Single Rigid Body Model (SRB)}
The quadruped system shown in Fig.~\ref{fig:SRB} is modeled using SRB model \cite{di2018dynamic}

\begin{align} 
 \mathbf{\ddot{p}} &= \frac{ \mathbf{R} (\alpha_{i}\mathbf{F}_{i} + \alpha_{j}\mathbf{F}_{j})}{m} - \mathbf{g}  \label{eqn:SRB1} \\
 \ddot{\theta} &= \frac{ \mathbf{r}_{i} \times\alpha_{i}\mathbf{F}_{i} + \mathbf{r}_{j} \times\alpha_{j}\mathbf{F}_{j}}{I} \label{eqn:SRB2}
\end{align}
where 
$\mathbf{p} \in \mathbb{R}^{2}$  and $ \mathbf{\dot{p}} \in  \mathbb{R}^{2}$  are the SRB center of mass (CoM) position and velocity both in the world frame respectively; $\theta$ is the pitch angle in world frame, and $\dot{\theta}$ is pitch angular velocities in body frame. The leg ground reaction force for the foot $i,j$ is $ \mathbf{F}_{i,j} \in \mathbb{R}^{2} $ and the distance from the foot $i,j$ to the center of mass is $ \mathbf{r}_{i,j}  \in \mathbb{R}^{2} $;
$\alpha_{i}, \alpha_{j}$ are the scalars to do the model selection, where $\alpha_{i}=1, \alpha_{j}=1$ for trotting model with front and rear foots contact on the ground, $\alpha_{i}=0, \alpha_{j}=2$ for rear stance bounding model with only rear foots contact on the ground,  $\alpha_{i}=2, \alpha_{j}=0$ for front stance bounding model with only front foot contact on the ground, and $\alpha_{i}=0, \alpha_{j}=0$ for flight phase;
 $\mathbf{R} \in \mathbb{R}^{2 \times 2}$ is the rotation matrix mapping vector from body frame to global frame; $m$ is the mass of the robot; $ \mathbf{g} \in  \mathbb{R}^{2}$ is the gravity vector in the world frame; $I $ is the inertia  of the torso in the body frame. Note that Eqs.~\ref{eqn:SRB1}-\ref{eqn:SRB2} are nonlinear and they can be compactly written as: $\dot{{\bf x}} = {\bf f}({\bf x}) + {\bf g}({\bf x}) {\bf u}$,  where $\mathbf{x}=[\mathbf{p},    \theta,      \mathbf{\dot{p}},   \dot{\theta}   ]^\top $ and ${\bf u} = [{\bf F}_i,{\bf F}_j]^\top$.

\subsection{Koopman Operator Theory}

For a given nonlinear system $ \mathbf{x}_{t+1}= \mathbf{f}(\mathbf{x}_{t}, \mathbf{u}_{t} ) ; \ \mathbf{x} \in \mathcal{X} \subseteq \mathbb{R}^n; \ \mathbf{u} \in \mathcal{U} \subseteq \mathbb{R}^m; \ \mathbf{f} : \mathcal{X} \times \mathcal{U} \to \mathcal{X} $, a set of nonlinear observable functions  $ \mathbf{\Pi}(\mathbf{x}) $ exists such that the evolution of the system along these observables is characterized by linear dynamics governed by an infinite dimension operator $ \mathbf{\mathcal{K}} $, known as Koopman operator \cite{koopman1931hamiltonian}
\begin{align}
    [\mathbf{\mathcal{K}}\mathbf{\Pi}](\mathbf{x},\mathbf{u})= \mathbf{\Pi} \circ \mathbf{f}( \mathbf{x}, \mathbf{u} ) 
\end{align}
A finite-dimensional approximation of $ \mathbf{\mathcal{K}}  $, denoted as $ \mathbf{K} = [\mathbf{A}, \mathbf{B}];  \mathbf{A} \in \mathbb{R}^{N \times N};  \mathbf{B} \in \mathbb{R}^{N \times m} $, is derived by employing the Extended Dynamic Mode Decomposition (EDMD) approach \cite{williams2015data}, which projects $ \mathbf{\mathcal{K}} $ onto a subspace of  observable functions via least squares regression. The finite dimension approximated operator $ \mathbf{K} $ can be used to represent the linear evolution of observable functions as

\begin{align}
    \mathbf{\Pi}(\mathbf{x}_{t+1}) = \begin{bmatrix} \mathbf{A}& \mathbf{B} \end{bmatrix} \begin{bmatrix} \mathbf{\Pi}(\mathbf{x}_{t}) \\ \mathbf{u}_{t} \end{bmatrix} = \mathbf{K}\mathbf{\hat{\Pi}}(\mathbf{x}_{t}, \mathbf{u}_{t}) \label{eqn:KP}
\end{align}
where $\mathbf{\Pi}(\mathbf{x}_{t}) = [\pi_{1}(\mathbf{x}_{t}),...,\pi_{N}(\mathbf{x}_{t})]^T \in \mathbb{R}^{N}$ is the dictionary of observable functions. By utilizing $M$ snapshots of the system states and control inputs in lifted space formed by the basis function $ \pi_i(\mathbf{x}) $ in the dictionary, we obtain the approximated operator $ \mathbf{K} $. The paired dataset $ \mathbf{X}=[\mathbf{x}_{1},\mathbf{x}_{2},...,\mathbf{x}_{M-1}]$ and $ \mathbf{Y} = [\mathbf{x}_{2},\mathbf{x}_{3},...,\mathbf{x}_{M}] $ can be obtained by perturbing nonlinear dynamics $ \mathbf{x}_{t+1}=\mathbf{f}(\mathbf{x}_{t}, \mathbf{u}_{t})  $ with a given control sequence $ \mathbf{U}=[\mathbf{u}_{1},\mathbf{u}_{2},...,\mathbf{u}_{M-1}] $, then the approximation of the Koopman operator  can be obtained by solving the least square regression \cite{li2017extended} such that 

\begin{align}
\mathbf{K} = \arg\min_{\mathbf{K}} \|  \mathbf{\Pi}(\mathbf{Y}) - \mathbf{K} 
 \mathbf{\hat{\Pi}}(\mathbf{X, U})   \|^{2}
\end{align}
By constructing $ \mathbf{G}_{1} $ and $ \mathbf{G}_{2} $, an analytical solution for $\mathbf{K}$ may be computed
\begin{align}
    \mathbf{G}_{1} &= \frac{1}{M} \sum_{i=1}^{M} \mathbf{\Pi}(\mathbf{y}_i) \hat{\mathbf{\Pi}}(\mathbf{x}_i, \mathbf{u}_{i})^\top \\
\mathbf{G}_{2} &= \frac{1}{M} \sum_{i=1}^{M} \mathbf{\Pi}(\mathbf{x}_i) \hat{\mathbf{\Pi}}(\mathbf{x}_i, \mathbf{u}_{i})^\top \\
\mathbf{K} &=\mathbf{G}_{1}\mathbf{G}_{2}^{-1}
\end{align}

\subsection{Koopman Operator-Based Modeling} \label{sec:KP_model}
The 2D SRB model equations (see Eqs.~\ref{eqn:SRB1}-\ref{eqn:SRB2}) are nonlinear. Our goal is to compute a linear model using Koopman operator theory.

We identify a set of Koopman observer functions  $ \mathbf{\Pi}(\mathbf{x}): \mathbb{R}^{n} \rightarrow \mathbb{R}^{N } $ that can evolve linearly in the lifted observable space with the finite-dimensional Koopman operator $ \mathbf{K}=[\mathbf{A}, \mathbf{B}] $, such that the dynamics are approximated by Eq.~\ref{eqn:KP}. To perform the EDMD to find the linear Koopman predictor $ \mathbf{A} \in \mathbb{R}^{N \times N}$ and  $\mathbf{B} \in \mathbb{R}^{N \times m} $ , a set of physics informed observable functions \cite{narayanan2023se} $ \bar{\mathbf{\Pi}} = [\underline{\mathbf{R}\dot{\theta}}, \underline{\mathbf{R}\dot{\theta}^{2}}, ..., \underline{\mathbf{R}\dot{\theta}^{p}}]^\top $ is selected to form the lifted state space, 
 where the operator $ \underline{(\cdot)} : \mathbb{R}^{l \times l} \rightarrow \mathbb{R}^{l^2 }$ maps the matrix into a vector by concatenating the columns inside the matrix, then the linear SRB states can be augmented as
\begin{align}
    \mathbf{\Pi} = [1, \mathbf{p},    \theta,      \mathbf{\dot{p}},   \dot{\theta}, \bar{\mathbf{\Pi}}]^\top \in   \mathbb{R}^{7+4p}
\end{align}
Note that one of the observable is $1$ here ,which is needed to include constant terms for the gravity. We also include the state $\mathbf{x}=[\mathbf{p},    \theta,      \mathbf{\dot{p}},   \dot{\theta}   ]^\top $ in the dictionary so we can recover it for its usage in the LMPC discussed in the next section.

\subsection{Koopman Operator-Based Model Predictive Control} \label{sec:KP_MPC}
 Unlike \cite{di2018dynamic} \cite{ding2019real}, which use MPC formulation to plan the flight phase, our approach, inspired by Raibert's controller \cite{raibert1990trotting}, performs all calculations during the stance phase.
We formulate a LMPC to track reference trajectories $ \mathbf{x}^{d}_{t} $ using the Koopman operator model as follows 
\begin{align} 
&\min_{\mathbf{u}}\sum_{i=0}^{k-1}\left\| \mathbf{x}_{t+i}-\mathbf{x}^{d}_{t} \right\|_{\mathbf{Q}_{i}} + \left\| \mathbf{u}_{t+i} \right\|_{\mathbf{R}_{i}} \label{eqn:MPC_cost} \\
 \text{s.t.} &\quad \mathbf{\Pi}_{t+i} = \mathbf{A}\mathbf{\Pi}_{i} + \mathbf{B}\mathbf{u}_{i}, \: i=0,1...k-1  \label{eqn:MPC_model1} \\
 &\quad \mathbf{x}_{i}= {\bf C}_x\mathbf{\Pi}(\mathbf{x}_{i}), \: {\bf u}_{\mbox{\scriptsize min}} \leq {\bf u}_i \leq {\bf u}_{\mbox{\scriptsize max}}   \label{eqn:MPC_control3} 
\end{align}
where ${\bf C}_x \in \mathbb{R}^{n \times N} $ is selection matrix that pulls out the state ${\bf x}$ from ${\bf \Pi(\mathbf{x})}$; 
$ \mathbf{Q}_{i} \in \mathbb{R}^{n \times n} $ and $ \mathbf{R}_{i} \in  \mathbb{R}^{m \times m} $ are user-chosen diagonal positive definite matrices.

For a given initial state ${\bf x}_0 \in \mathbb{R}^{n} $, using Eqs.~\ref{eqn:MPC_model1} and \ref{eqn:MPC_control3} recursively for $k$ time steps, we obtain 
\begin{align}
    \mathbf{X}_{qp} = \mathbf{A}_{qp}\mathbf{x}_{0} + \mathbf{B}_{qp}\mathbf{U}_{qp}
\end{align}
where $\mathbf{X}_{qp} \in  \mathbb{R}^{n \times k} $ and $ \mathbf{U}_{qp} \in \mathbb{R}^{m \times k} $  are the concatenated state and control from $1,2,...,k$.
The cost function can then be rewritten as
\begin{align}
    &\min_{\mathbf{U}_{qp}}\left\| \mathbf{A}_{qp}\mathbf{x}_{0} + \mathbf{B}_{qp}\mathbf{U}_{qp}-\mathbf{X}^{d}_{qp} \right\|_{\mathbf{Q}_{qp}} + \left\| \mathbf{U}_{qp} \right\|_{\mathbf{R}_{qp}}
\end{align}
where $ \mathbf{X}^{d}_{qp} \in    \mathbb{R}^{n \times k}$ is the concatenated reference trajectories; \(\mathbf{Q}_{qp} \in \mathbb{R}^{nk \times nk} \) and \(\mathbf{R}_{qp} \in  \mathbb{R}^{mk \times mk} \) are user-chosen diagonal positive weight matrix.

The QP can now be written as
\begin{align}
    &\min_{\mathbf{U}_{qp}} \quad \frac{1}{2}\mathbf{U}_{qp}^\top \mathbf{H} \mathbf{U}_{qp} + \mathbf{P} \mathbf{U}_{qp} \\
    \text{s.t.} &\quad \mathbf{\underline{c}}  \leq \mathbf{C}\mathbf{U}_{qp} \leq  \mathbf{\overline{c}} 
\end{align}
where $\mathbf{H} = 2(\mathbf{B}_{qp}^\top \mathbf{Q}_{qp} \mathbf{B}_{qp}+ \mathbf{R}_{qp})$, $\mathbf{P} = 2(\mathbf{x}_{0}^\top \mathbf{A}_{qp}^\top \mathbf{Q}_{qp} \mathbf{B}_{qp} - \mathbf{X}_{qp}^{d\mathstrut\top}  \mathbf{Q}_{qp} \mathbf{B}_{qp})$ and $\mathbf{C} \in \mathbb{R}^{mk \times mk}, \mathbf{\underline{c}} \in \mathbb{R}^{mk}, \mathbf{\overline{c}} \in \mathbb{R}^{mk}$ denote the inequality constraints of control inputs.

\subsection{Controller Implementation}
The controller is implemented on a model of Unitree Go1 quadruped in MuJoCo version 2.0.0 \cite{todorov2012mujoco} using Ubuntu 20.04 on an Intel Core i7 machine.

\subsubsection{Finite State Machine}
A finite state machine is used to manage the control logic, sending commands  to leg controllers for FR, FL, RR, and RL legs. In our trot gait controller, FR-RL and FL-RR legs move in pairs, transitioning between swing and stance phases every $0.2-0.5$ sec. If a swing leg collides with an obstacle, detected by a contact sensor, the swing foot's position is held constant. In our bounding gait controller, FR-FL and RR-RL legs move in pairs, transitioning between front stance, flight phase, rear stance, and flight phase following a duty cycle where flight time, front stance time and rear stance time are $0.1$ sec, $0.1$ sec, $0.05$ sec respectively. If legs collide with an obstacle during the flight phase, detected by a contact sensor, the foot's position is held constant. 

\subsubsection{Swing/Flight Phase Controller}
The role of the swing/flight leg controller is to track the reference position of the foot in these two phases using joint torques. Given the foot reference position and velocity, $\mathbf{p}_{fi}^{d},\mathbf{\dot{p}}_{fi}^{d} $,  an analytical inverse (see \cite{robotics12020035} Sec. 3.5) is used to compute the corresponding joint reference position and velocity, $ \mathbf{q}_{i}^{d} ,\mathbf{\dot{q}}_{i}^{d} $.  The following simple proportional-derivative controller is used to compute the leg $i$ joints torques.
\begin{align}
    \mathbf{\tau}_{i} = -\mathbf{K}_{p}(\mathbf{q}_{i} - \mathbf{q}_{i}^{d})  - \mathbf{K}_{d}(\mathbf{\dot{q}}_{i} - \mathbf{\dot{q}}_{i}^{d} )
\end{align}

\subsubsection{Linear Model Predictive Control}
The LMPC uses the Koopman operator-based model and estimated torso states to compute desired ground reaction forces for the stance legs (see Sec.~\ref{sec:KP_MPC}). The planning horizon for the Model Predictive Control is $6$ ms or $166.67$ Hz while the update horizon is $5$ ms or $200$ Hz. The LMPC is solved online using qpSWIFT \cite{pandala2019qpswift} in about $3$ ms. 

\subsubsection{Stance Phase Controller}
The stance leg controller module uses the desired ground reaction forces to estimate the joint torques using the Jacobian of the stance leg. \begin{align}
    \mathbf{\tau}_{i} = \mathbf{J}_{i}^\top{} \mathbf{f}_{i}
\end{align}

\begin{figure} [tbp]
\includegraphics[scale=1]{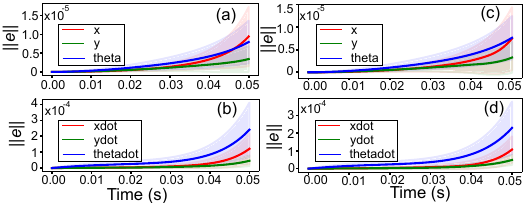}
\caption{2D Single Rigid Body model Koopman operator fitting result}
\label{fig:koopman_fit}
\end{figure}

\section{Results}

\subsection{Koopman Operator Model Fit}

To generate data for the EDMD, we integrate the SRB Eqs.~\ref{eqn:SRB1}-\ref{eqn:SRB2} with Runge-Kutta of order 4 with a fixed step size of $dt=0.001$ sec from $t=0$ to $t=0.1$ sec. For each roll-out, we use a randomly generated initial conditions at $t=0$ and random control input for every $0.001$ sec. We use a total of $100$ roll-outs to create a training dataset. Then using the $p=4$ observables ${\bf \Pi}({\bf x})$ discussed in Sec.~\ref{sec:KP_model}, we perform the EDMD to obtain a linear model.

 Fig.~\ref{fig:koopman_fit} (a)-(b), (c)-(d) shows the fitting result for trotting model and rear legs bounding models respectively. We did the extensive test by generating  $50$ initial conditions and using $50$ random force profiles for a $0.05$ sec roll-out using the SRB model, then using the same paired initial condition and force inputs we generated corresponding predictions using Koopman operator model. The linear model fitting results are evaluated by the error between the actual and predicted trajectories across 50 sets of data  shown in Fig.~\ref{fig:koopman_fit}, 
 where (a) (c) for translation and orientation $\mathbf{p}, \theta$ states, (b) (d) for linear velocity and angular rates ${ \mathbf{\dot{p}}, \dot{\theta} }$ states. The solid lines show the mean and the bands show the variance. It can be seen that the errors are within $\pm 10^{-4}$ indicating a sufficiently accurate fit.

\begin{figure} [tbp]
\includegraphics[scale=0.95]{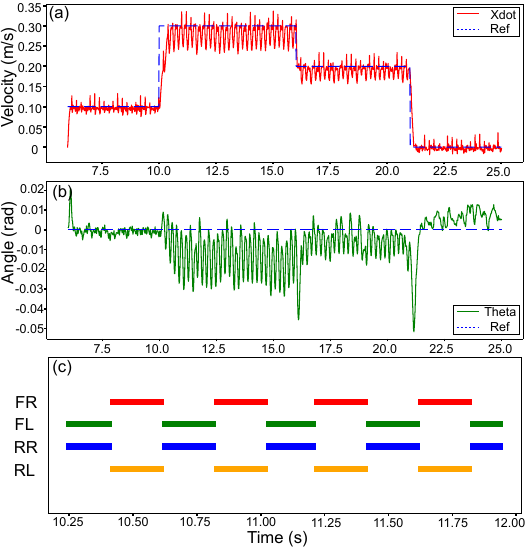}
\caption{Trotting forward velocity tracking  }
\label{fig:trot_step}
\end{figure}

\subsection{Koopman Operator LMPC in Trotting}
To check the trotting gait reference tracking ability, we command the robot to follow a set of reference trajectory as shown in Fig.~\ref{fig:trot_step} (a). The reference is parameterized by a step function, starting at $6$ sec. The forward velocity is commanded to change sequentially: first to $0.1$ m/s, then $0.3$ m/s, followed by $0.2$ m/s, and finally back to $0$ m/s. At $10$ sec, the command velocity undergoes a sharp increase from $0.1$ m/s to $0.3$ m/s, yet the controller successfully achieves stable forward velocity tracking with overall RMSE as $0.029$. From Fig.~\ref{fig:trot_step} (b), one can see that when the forward reference velocity is relatively high, the controller temporarily allows the robot to pitch forward at a small angle to improve forward tracking performance. When the commanded forward velocity is set to zero, the pitch also returns to its neutral position. The trotting gait pattern diagram is shown in Fig.~\ref{fig:trot_step} (c) where the bars indicate the period of ground contact.

A rough terrain shown in Fig.~\ref{fig:cover_pic} is designed to test the controller's disturbance rejection capability. The terrain is scattered with eight blocks, cubes, and capsules, ranging in height from $5$ to $7$ cm, placed on the path in front of the quadruped, which is commanded to track a forward velocity of $0.3$ m/s. The experimental results are shown in Fig.~\ref{fig:gait_terrain} (a) (b). Around $3$ sec and $9$ sec, two slip incidents occur, causing the quadruped to experience a severe forward pitch. However, the controller successfully restores the robot's balance. Between $3$ and $8$ sec, the velocity tracking deviates from the reference due to the rough terrain, but the robot manages to maintain pitch stability, with a tracking error within $\pm3$ degrees while traversing.

\begin{figure} [tbp]
\includegraphics[scale=0.95]{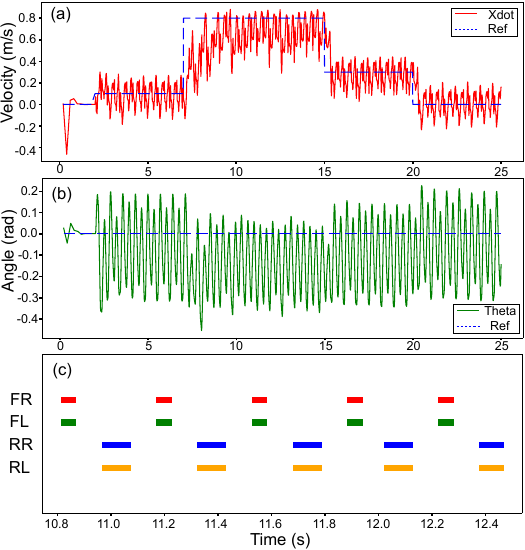}
\caption{Bounding forward velocity tracking}
\label{fig:bound_step}
\end{figure}

\subsection{Koopman Operator LMPC in Bounding}
To check the trotting gait reference tracking ability, we command the robot to follow a combination of reference forward 
trajectory as shown in Fig.~\ref{fig:bound_step} (a). Compared to trotting, bounding allows the robot to achieve higher forward velocity tracking due to its higher gait frequency and additional flight times. Starting at $2$ sec, the reference signal is defined by a step function that sequentially commands changes in forward velocity. The velocity first increases to $0.1$ m/s, then jumps to $0.8$ m/s, decreases to $0.3$ m/s, and finally returns to $0$ m/s. Notably, at $7$ sec, there's a sharp increase from $0.1$ m/s to $0.8$ m/s in the commanded velocity. Despite this sudden change, the controller successfully maintains stable forward velocity tracking with overall RMSE $ 0.189$. Similar to the trotting gait, the same pitch angle pattern is observed in the bounding gait. From $7$ to $15$ sec, the forward velocity is commanded to be relatively high. To achieve effective forward tracking, the controller allows the pitch angle to tilt forward, and as the reference forward velocity decreases, the pitch angle returns to its neutral position. The bounding gait pattern diagram is shown in Fig.~\ref{fig:bound_step} (c) where the bars indicate the period of ground contact.

\begin{figure} [tbp]
\includegraphics[scale=0.95]{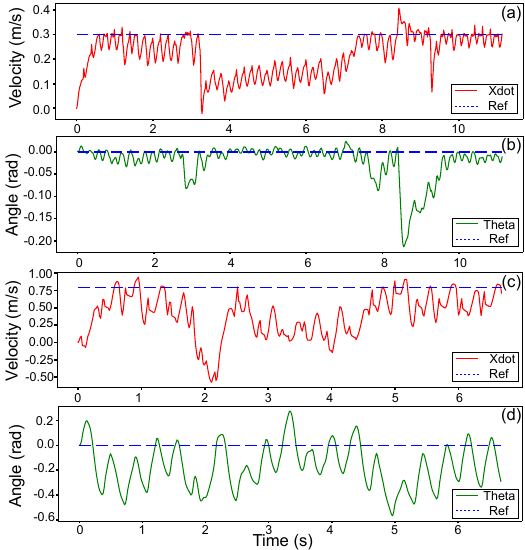}
\caption{Trotting (top two) and bounding (bottom two) on rough terrain}
\label{fig:gait_terrain}
\end{figure}

A rough terrain profile is  used to test the disturbance rejection capability of the bounding gait. The experimental results are shown in Fig.~\ref{fig:gait_terrain} (c) (d). In the bounding experiment, the robot is commanded to move at a forward velocity of $0.75$ m/s shown in Fig.~\ref{fig:gait_terrain} (c). Around $2$ sec, the front foot stumbles on a block, causing a sharp decrease in forward velocity. Between $2$ and $5$ sec, the forward velocity tracking performance deviates due to the rough terrain; however, the controller is able to maintain a stable, periodic bounding pitch angle, as shown in Fig.~\ref{fig:gait_terrain} (d).

\subsection{Koopman Operator LMPC in Gait Switch}
\begin{figure} [tbp]
\includegraphics[scale=0.95]{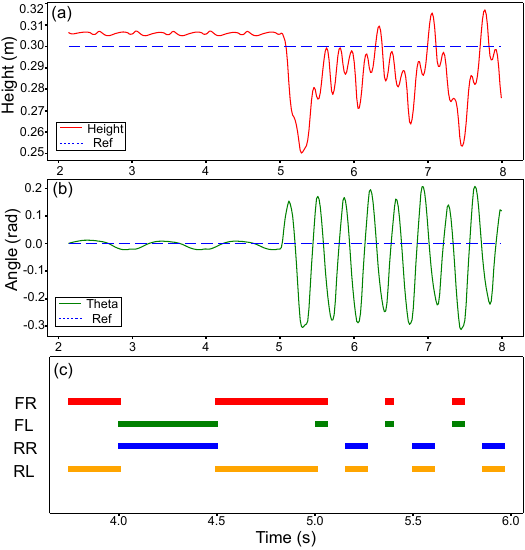}
\caption{Trotting transit to bounding}
\label{fig:trot_2_bound}
\end{figure}

\begin{figure} [tbp]
\includegraphics[scale=0.95]{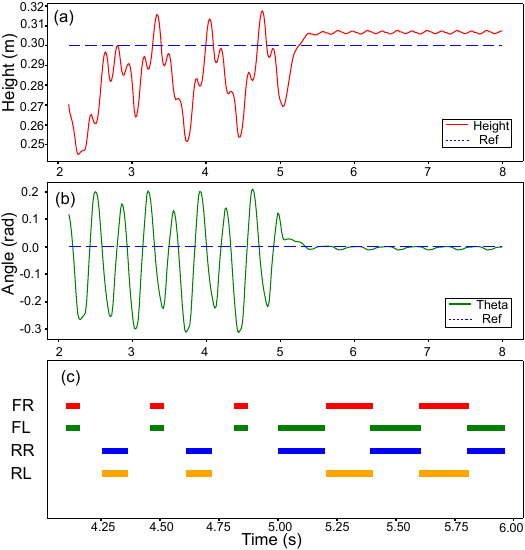}
\caption{Bounding transit to trotting }
\label{fig:bound_2_trot}
\end{figure}

In the gait switch experiment, two transitions are conducted: one from trotting to bounding, and the other from bounding to trotting, with the results shown in Fig.~\ref{fig:trot_2_bound}  and Fig.~\ref{fig:bound_2_trot} respectively. The gait transitions occur while the robot is trotting or bounding in place. The two quadruped gaits are governed by separate finite state machines, and the transitions are based on a fixed switch time. Specifically, the transition from trot to bound and from bound to trot lasted $6$ sec, with the first $3$ sec consisting of one gait and the next $3$ sec consisting of the other.

For the transition from trotting to bounding, the switch time can be placed at any point along the time axis due to the nature of the trotting gait, which always has a stance leg on the ground, providing better support polygon. On the other hand, one needs to choose the switch more carefully when transitioning from bounding to trotting. Due to the distinct phase composition of bounding, a large pitch moment may be induced if the switch occurs during bounding stance phase. For instance, if the switch happens during the front stance phase while the back foot is in the air, switching to the trotting gait will cause the back foot to kick the ground, generating a significant impulse that can disturb the robot's balance. Inspired by \cite{raibert1990trotting}, we placed our bounding to trotting switch time in the fly phase to realize the smooth gait transition.

\section{Discussion, Conclusion, and Future Work}
In this paper, we have used the Koopman operator to create a linear model of the 2D SRB model of a quadruped. This is used with LMPC to control the quadruped trotting, bounding gaits and theirs transition demonstrating the efficacy of the approach.




  We do see that our linear model remains accurate with a prediction error within $ \pm 10^{-4} $ up to $0.05$ sec, its accuracy diminishes as time progresses. 
This limitation suggests that the linear model may not fully capture the complexities of the system over longer duration. To address this issue, one could use a neural network as a basis function \cite{yeung2019learning}, which would provide a more flexible and accurate fit.
Alternatively, a bilinear Koopman operator-based model could provide a better representation of the system's dynamics, due to the SRB control affine feature \cite{bruder2021advantages}. But the use of bilinear model leads to a Nonlinear Model Predictive Control which is computationally challenging to solve \cite{folkestad2021koopman}. One approach is to linearize the bilinear term at the operator point, which leads to an LMPC \cite{yu2022autonomous}.


In this work, we implement simple gaits. Since most gaits are inherently periodic, the gait frequency can be easily modified by changing the cycle duration \cite{winkler2018gait}, while gait speed can be adjusted by altering the phase combination and frequency. For example, a flying trot can increase forward velocity by incorporating a flight phase within the duty cycle \cite{di2018dynamic}.

Our future work will explore methods to increase the robustness of the quadruped to larger disturbances, increase the range of movement of the robot, and using a neural network as the basis function to increase the  model accuracy.

\bibliographystyle{IEEEtran}
\bibliography{pranav_bib2}

\end{document}